\documentclass[conference]{IEEEtran}
\IEEEoverridecommandlockouts

\usepackage[table,xcdraw]{xcolor}
\usepackage{graphicx}
\usepackage{amsmath}
\usepackage{amssymb}
\usepackage{multirow}
\usepackage{xcolor}
\usepackage{colortbl}
\usepackage{pgfplots}
\usepackage{layout}
\usepackage{amssymb}
\usepackage{bbding}
\usepackage{booktabs}
\usepackage{orcidlink}
\usepackage{diagbox}

\usepackage{stfloats}
\usepackage[misc]{ifsym}
\usepackage{makecell}
\usepackage{caption}
\usepackage{ulem}

\usepackage[T1]{fontenc}

\usepackage{graphicx,verbatim}
\usepackage{orcidlink}
\usepackage{fancyhdr}
\usepackage{hyperref}

\def\BibTeX{{\rm B\kern-.05em{\sc i\kern-.025em b}\kern-.08em
    T\kern-.1667em\lower.7ex\hbox{E}\kern-.125emX}}
\begin{document}

\title{MM-UNet: Morph Mamba U-shaped Convolutional Networks for Retinal Vessel Segmentation
\\
}


\author{
\IEEEauthorblockN{1\textsuperscript{st}  Jiawen Liu$^\dagger$
\orcidlink{0009-0002-6827-6049}
}

	\IEEEauthorblockA{\textit{South China University of Technology} \\ GuangZhou, China \\
		202321044682@mail.scut.edu.cn}
	
    \and
	\IEEEauthorblockN{2\textsuperscript{nd} Yuanbo Zeng$^\dagger$}
	\IEEEauthorblockA{\textit{Jiangxi University of Finance and Economics} \\ Nanchang, China \\
		2202320786@stu.jxufe.edu.cn }
	
    \and
	\IEEEauthorblockN{3\textsuperscript{rd} Jiaming Liang$^\dagger$ \orcidlink{0009-0000-6007-9759}
    }
	\IEEEauthorblockA{\textit{South China University of Technology} \\ GuangZhou, China \\
		csliangjm@mail.scut.edu.cn}
	
    \and
        
	\IEEEauthorblockN{4\textsuperscript{th}Yizhen Yang}
	\IEEEauthorblockA{\textit{New York University} \\ New York, USA \\
		yy4304@nyu.edu}
	
	\and
                \IEEEauthorblockN{5\textsuperscript{th}	Yiheng Zhang}
	\IEEEauthorblockA{\textit{Sichuan University} \\ Chengdu , China\\
		zhangyiheng.nov@gmail.com}
	
	\and
	\IEEEauthorblockN{6\textsuperscript{th} Enhui Cai}
	\IEEEauthorblockA{\textit{South China University of Technology} \\ GuangZhou, China \\
		enhuicai28@gmail.com}
	
    \and
	\IEEEauthorblockN{7\textsuperscript{th} Xiaoqi Sheng \orcidlink{0000-0002-2929-5805}
    }
	\IEEEauthorblockA{\textit{South China University of Technology} \\ GuangZhou, China \\
		xqsheng@scut.edu.cn}
	
    \and
        \IEEEauthorblockN{8\textsuperscript{th} Hongmin Cai$^{*}$ \orcidlink{0000-0002-2747-7234}
        }
	\IEEEauthorblockA{\textit{South China University of Technology} \\ GuangZhou, China \\
		hmcai@scut.edu.cn}
        
}
\maketitle
\thispagestyle{fancy}
\fancyhead{}
\lfoot{
\footnotesize
$^{\dagger}$: These authors contributed equally to this work.\\
* Corresponding author: \href{https://www2.scut.edu.cn/bioinformatics/sysPIjs/list.htm}{Hongmin Cai} (hmcai@scut.edu.cn).
 }
\cfoot{\quad}
\renewcommand{\headrulewidth}{0pt}
      \renewcommand{\footrulewidth}{0pt}

\begin{abstract}
Accurate detection of retinal vessels plays a critical role in reflecting a wide range of health status indicators in the clinical diagnosis of ocular diseases. Recently, advances in deep learning have led to a surge in retinal vessel segmentation methods, which have significantly contributed to the quantitative analysis of vascular morphology. 
However, retinal vasculature differs significantly from conventional segmentation targets in that it consists of extremely thin and branching structures, whose global morphology varies greatly across images. These characteristics continue to pose challenges to segmentation precision and robustness.
To address these issues, we propose MM-UNet, a novel architecture tailored for efficient retinal vessel segmentation. The model incorporates Morph Mamba Convolution layers, which replace pointwise convolutions to enhance branching topological perception through morph, state-aware feature sampling. Additionally, Reverse Selective State Guidance modules integrate reverse guidance theory with state-space modeling to improve geometric boundary awareness and decoding efficiency. Extensive experiments conducted on two public retinal vessel segmentation datasets demonstrate the superior performance of the proposed method in segmentation accuracy. Compared to the existing approaches, MM-UNet achieves F1-score gains of 1.64 $\%$ on DRIVE and 1.25 $\%$ on STARE, demonstrating its effectiveness and advancement. The project code is public via 
\href{https://github.com/liujiawen-jpg/MM-UNet}{https://github.com/liujiawen-jpg/MM-UNet}.
\end{abstract}


\begin{figure*}[!t]
    \centering      
    \includegraphics[width=\textwidth]{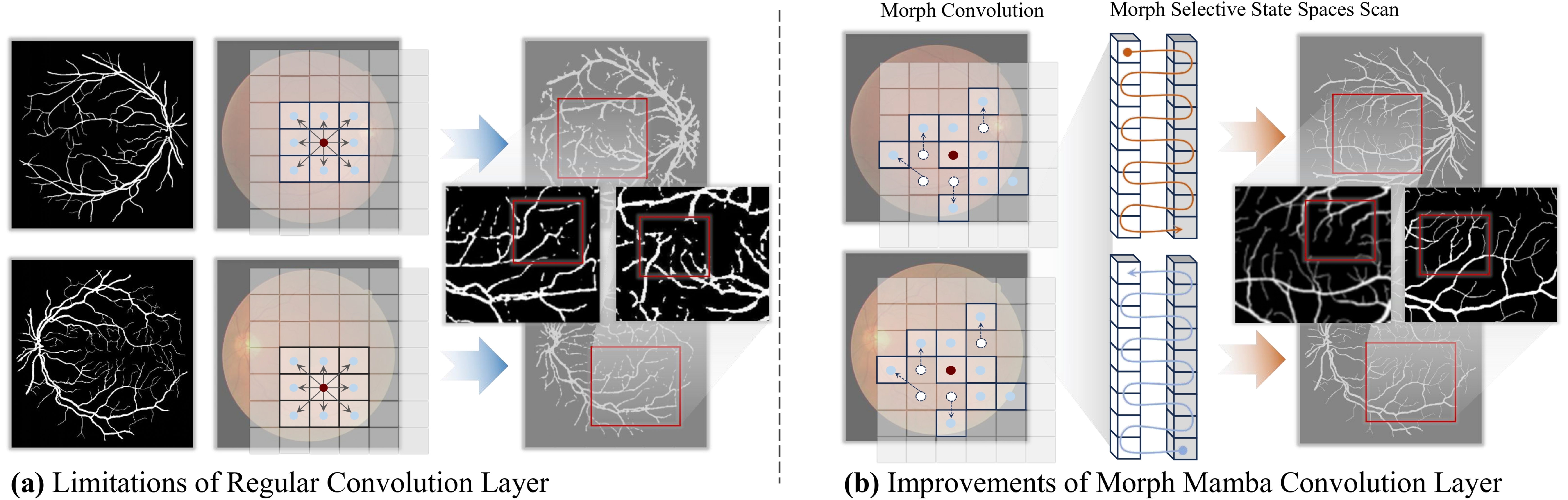}
    \caption{(a) \textbf{Limitations}: regular convolutional layers fail to accurately capture intricate vessel topology; (b) Improvements: MMC layers integrate a dynamic morph convolution mechanism with morph state-space modeling to effectively construct accurate topological representations.}
    \label{fig: fig1}
\end{figure*}

\section{Introduction}
In recent years, the growing demand for early diagnosis of retinal fundus diseases has drawn increasing attention to automated retinal vessel analysis~\cite{cen2021automatic}. With advances in imaging techniques and computational hardware, Deep Learning (DL) based methods~\cite{liang2025sfit, pang2025slim,liu2022full} have emerged as the dominant approach for Retinal Vessel Segmentation~(RVS). These methods offer high precision in vascular abnormalities, providing critical support for disease screening, diagnostic efficiency, and streamlined clinical workflows~\cite{wang2024advancing}.

Unlike lesion~\cite{liang2024comprehensive,liang2025hwa, zhou2023ga} or organ segmentation~\cite{tadokoro2023pre,pang2023slim}, retinal-vessel segmentation confronts a precision bottleneck~\cite{qin2024review, liang2022u}.  
The main culprit is the vessels’ highly intricate tubular topology and complex clinical imaging.
As shown in Fig.~\ref{fig: fig1} (a), retinal vessels possess extremely thin terminal branches and display significant global morphological deformation. This results in vessel targets occupying only a small portion of the local pixel space in retinal images, which poses a serious challenge for deep learning models built on convolutional operations. 
Those models, such as U-Net~\cite{ronneberger2015u} and its variants~\cite{dai2024i2u,liang2023agilenet}, rely heavily on standard convolutional upsampling operations,  limiting their ability to precisely capture high-resolution peripheral details and frequently leading to fragmented or disconnected segmentation outputs.  
With the emergence of Vision Transformer models in the field of retinal vessel segmentation~\cite{tan2024deep,kreitner2024synthetic,pang2023slim}, global feature extraction capabilities have been partially improved. However, these methods still lack the capacity to provide precise guidance for capturing delicate local patterns, particularly at vessel termini~\cite{qin2024review}. 
Recent methods such as DCASU-Net~\cite{xu2023dcsau}, FR-UNet~\cite{liu2022full}, and FSG-Net-L~\cite{seo2025full} have improved local representation using multi-scale fusion and high-resolution strategies, yet accurately capturing the full vascular topology remains challenging.

To tackle the above obstacles, this work proposes MM-UNet (\textbf{M}orph \textbf{M}amba \textbf{U}-shaped convolutional \textbf{Net}works), a novel segmentation framework featuring two innovative components: MMC (\textbf{M}orph \textbf{M}amba \textbf{C}onvolution) layers and RSSG (\textbf{R}everse \textbf{S}elective \textbf{S}tate \textbf{G}uidance) module.
Specifically, as shown in Fig.~\ref{fig: fig1} (b), the MMC layers tackle the intricate topological tubular structure of retinal vessels by integrating a dynamic morph convolution mechanism with morph state-space modeling. This integration enables the adaptive capture of narrow and tortuous local features inherent in tubular structures, significantly enhancing geometric perception. By superseding pointwise convolutional layers within all of the U-shaped segmentation networks, MMC significantly enhances the model’s capability for efficient topological representation and modeling.
Concurrently, RSSG modules are introduced into the inter-level skip connections of the U-shaped architecture to provide additional geometric structural guidance during the upsampling stage. Through the combination of reverse guidance mechanisms and state-space modeling, RSSG modules extract complementary boundary information from both interior and exterior regions encoded during downsampling, effectively enhancing the model’s capacity to discern structural contours and maintain spatial coherence. 
Extensive experiments conducted on two widely used retinal image datasets, STARE and DRIVE, demonstrate the superior performance of MM-UNet and its key components compared to current state-of-the-art methods, achieving improvements of at least 1.64\% and 1.25\% in terms of F1-score, respectively.
In summary, our contributions are as follows:These authors contributed equally to this work
\begin{itemize}
        \item \underline{\textbf{\textit{Innovation.}}}
To tackle the challenges of fine and branching vascular structures, Morph Mamba Convolution is introduced, which is a novel state-aware morph sampling mechanism that significantly enhances topological perception.

  \item\underline{\textbf{\textit{Framework.}}}
A novel retinal vessel segmentation framework, MM-UNet, is proposed to address the unique challenges of fine, branching vascular structures. It replaces traditional pointwise convolutions with Morph Mamba convolution layers and integrates a high-performance Reverse Selective State Guidance module, thereby significantly enhancing its perception and delineation of geometric boundaries.

  \item\underline{\textbf{\textit{Validation.}}}
Extensive evaluations on diverse benchmark datasets validate the superiority of our method over state-of-the-art retinal vessel segmentation frameworks, revealing its strong generalization ability across varying image conditions and its effectiveness in reliably identifying complex and clinically relevant vascular structures.

    \end{itemize}

\section{Relate Works}
Deep learning-based RVS has witnessed substantial progress, with existing methods broadly categorized into four families: (1) Convolutional Segmentation Networks, (2) Graph-Based and Multi-Scale Hybrid Models, (3) Transformer and Attention-Enhanced Architectures, and (4) State-Space or Dual-Decoder Frameworks.

\textbf{Convolutional Segmentation Networks} mark the earliest deep learning solutions in this domain. U-Net~\cite{ronneberger2015u} pioneered the encoder–decoder paradigm with skip connections, enabling effective recovery of spatial details. However, standard convolutional methods often struggle with preserving the global continuity of thin, tortuous vessels, especially in regions with sparse signals.

\textbf{Graph-Based and Multi-Scale Hybrid Models} aim to better capture vessel topology and multi-scale feature representation. DE-DCGCN-EE~\cite{li2022dual} employs dynamic graph convolution and edge enhancement to model vessel connectivity, while GT-DLA-dsHFF~\cite{s22155826} integrates global and local attention via deep–shallow hierarchical fusion. BCU-Net~\cite{ZHANG2023106960} and PA-Net~\cite{luo2025pa} further combine multi-resolution encoding with adaptive fusion to improve detail recognition. Despite their structural modeling capabilities, these methods often exhibit limited boundary localization, leading to suboptimal F1 and sensitivity scores.

\textbf{Transformer and Attention-Enhanced Architectures} improve long-range dependency modeling, which is essential for maintaining vessel continuity. For example,
Wave‑Net~\cite{LIU2023106341} replaces U‑Net’s standard skip‑connections with a detail‑enhancement-and-denoising block to better preserve continuity even in ultra-thin vessel branches.
CoVi‑Net~\cite{JIANG2024108047} integrates a local‑global feature aggregation module and employs bidirectional weighted fusion to reinforce structural consistency across the segmented vasculature. 
However, this series of Transformer-based methods tends to suffer from overfitting during training and exhibits limited sensitivity to fine vascular branches~\cite{qin2024review}.


\textbf{State-Space or Dual-Decoder Frameworks} have recently emerged to balance modeling capacity with computational efficiency. MSPDD-Net~\cite{li2025mspdd} adopts dual decoders and wavelet edge enhancement, but its static context still limits local structural flexibility. HRD-Net~\cite{LIU2024108295} utilizes deformable convolutions in high-resolution pipelines to preserve microvascular details, yet may lack adaptability in sparse or irregular vessel regions. TP-Net~\cite{10018415} introduces a two-path design to decouple edge and trunk extraction, but may suffer from coarse-to-fine fusion inconsistencies.


\section{Methodology}
\begin{figure*}[!t]
    \centering      
    \includegraphics[width=\textwidth]{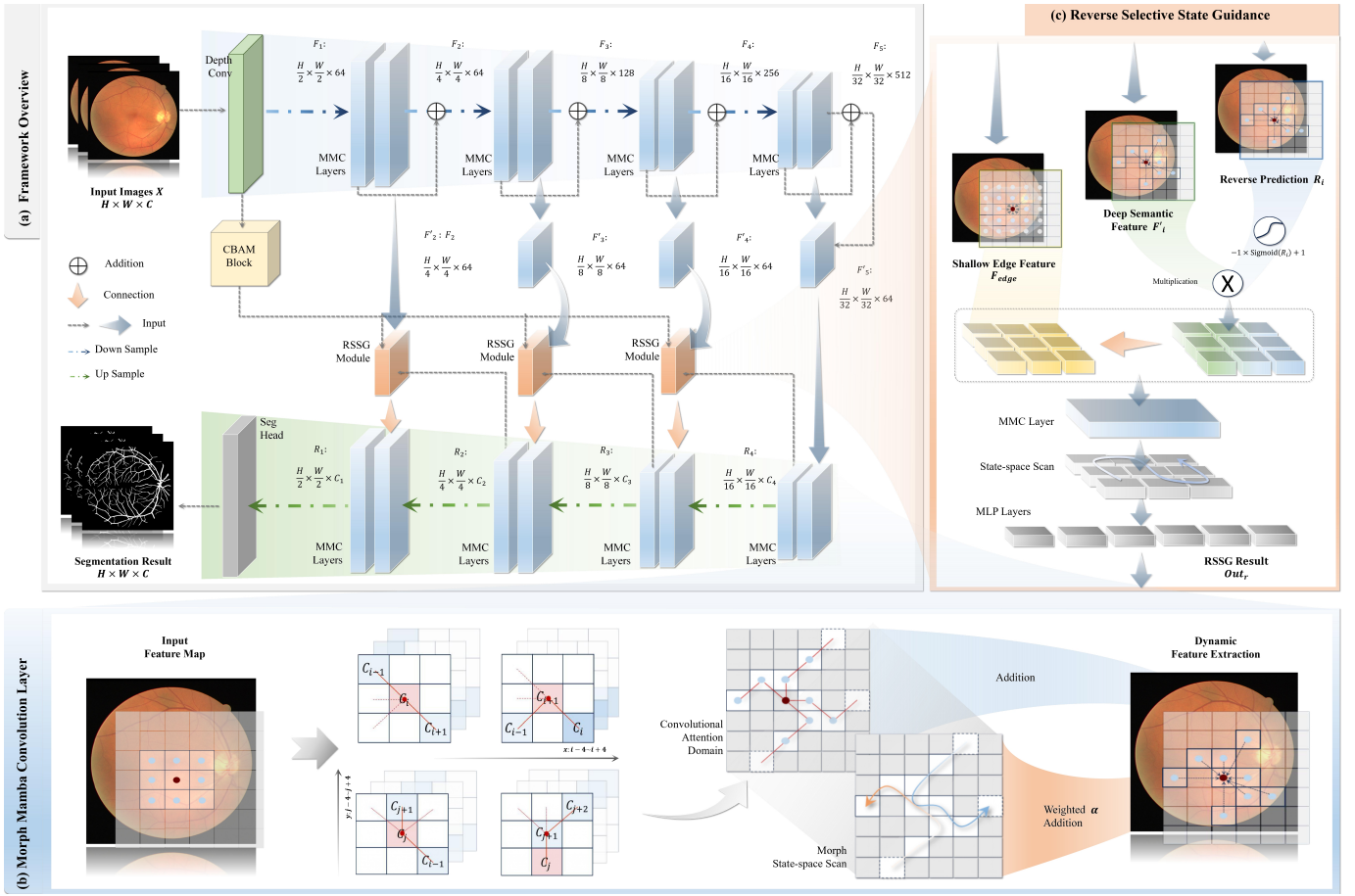}
    \caption{Overview of MM-UNet.}
    \label{fig: overview}
\end{figure*}

MM-UNet is proposed to tackle the challenge of complex topological structures in RVS, achieving high-performance segmentation via novel modular and architectural designs, as illustrated in Fig.~\ref{fig: overview} and detailed in subsequent sections.

\subsection{U-shaped Framework Overview}\label{se: frame}
MM-UNet adopts a U-shaped segmentation framework, which is shown in Fig.~\ref{fig: overview} (a), comprising a five-step encoder and a corresponding four-step decoder.  All pointwise convolutional layers are replaced by the proposed Morph Mamba Convolution (MMC) layers (detailed in Section~\ref{se: MMC}). The encoder is adapted from non-pretraining ResNet-34~\cite{he2016deep} to extract five downsampled features $F_i$ with dimensions $\frac{H}{2^i} \times \frac{W}{2^i} \times C_i$, where $C_1 = 64$ and $C_i = 64 \times 2^{i-2}$, $i \in [2, 3, 4, 5]$, from input images $x$ of size $H \times W \times 3$. To reduce computational cost~\cite{du2025net}, three MMC layers are used to unify the channels of $F_3$, $F_4$, and $F_5$ to 64 (defined as $F_3'$, $F_4'$, and $F_5'$). After background suppression by the Convolutional Block Attention Module (CBAM)~\cite{woo2018cbam}, the low-level feature $F_1$ is passed to the proposed Reverse Selective State Guidance (RSSG) modules (detailed in Section~\ref{se: RSSG}) and integrated into the decoder to guide upsampling and enhance geometric boundary awareness. Finally, multi-scale decoder features are fused and passed through a sigmoid layer to generate a pixel-wise probability map for accurate and scalable retinal vessel segmentation.

\subsection{Morph Mamba Convolution Layers}\label{se: MMC}
To overcome the limited sensitivity of traditional pointwise convolutions to complex branching topologies~\cite{yang2024d}, the MMC layer integrates morph convolution with a morph state-space computing mechanism, as illustrated in Fig.~\ref{fig: overview} (b).
Assume a conventional $3 \times 3$ kernel-based convolution layer with the convolution center located at $C_i = \{x_i, y_i\}$. Thus, the other coordinates within its receptive field can be expressed as in Eq.~\ref{eq:1a}. In MMC, a learnable morph offset $\Delta \in \{-1, 1\}$ is introduced to redefine the surrounding positions relative to the center, yielding $C_{i+c} = \{x_{i+c}, y_{i+c}\}$ and $C_{j+c} = \{x_{j+c}, y_{j+c}\}$, as shown in Eq.~\ref{eq:1b}, ~\ref{eq:1c}. Due to the fractional nature of $\Delta$, sub-pixel level displacements are allowed, and the corresponding feature values at non-integer positions are obtained via bilinear interpolation, as shown in Eq.~\ref{eq:1d}.

\begin{subequations}
\begin{align}
C = &{(x-1,y-1), (x-1,y), ..., (x+1,y+1)} \label{eq:1a} \\ 
C_{i \pm c} = &
\begin{cases}
(x_{i+c}, y_{i+c}) = (x_i + c,\, y_i + \sum\nolimits_i^{i+c} \Delta y) \\
(x_{i-c}, y_{i-c}) = (x_i - c,\, y_i + \sum\nolimits_i^{i-c} \Delta y)
\end{cases} \label{eq:1b} \\ 
C_{j \pm c} = &
\begin{cases}
(x_{j+c}, y_{j+c}) = \left( x_j + \sum\nolimits_{j}^{j+c} \Delta x,\, y_j + c \right) \\
(x_{j-c}, y_{j-c}) = \left( x_j + \sum\nolimits_{j-c}^{j} \Delta x,\, y_j - c \right)
\end{cases} \label{eq:1c} \\
C &= \sum_{C'} B(C',C) \cdot C' \label{eq:1d}
\end{align}
\end{subequations}
where $B(\cdot)$ is the bilinear weight kernel, $C$ denotes a fractional location for Eq.~\ref{eq:1b} and Eq. ~\ref{eq:1c}, $C'$ enumerates all integral spatial locations.

To enhance the model's perception of complex tubular structures, MMC incorporates a Multi-view Feature Fusion Strategy~\cite{qi2023dynamic} along with a selective state-space computation mechanism~\cite{gu2023mamba}, guiding the model to complement its attention to fundamental features from multiple perspectives. Specifically, for each position $C$, two feature maps $f^{l}(C_x)$ and $f^{l}(C_y)$ are extracted along the x-axis and y-axis, respectively, as computed in Eq.~\ref{eq:2a}. Meanwhile, based on a novel morph SSM computation, cross-axis contextual dependencies are constructed, as illustrated in Eq.~\ref{eq:2b}. Finally, as shown in Eq.~\ref{eq:2c}, the feature sampling operation is completed by stacking multi-view feature maps from different perspectives to form a correlated multi-view representation.

\begin{subequations}
\begin{align}
f^{l}(C) &= {\sum_{i} w(C_i) \cdot f^l(C_i), \sum_{j} w(C_j) \cdot f^l(C_j)}
\label{eq:2a} \\
f^{l}_m(C) &= \alpha \times Ma(F^{-1}(f^{l}(C)_{f})) + f^{l}(C)_{f} \label{eq:2b}\\
&T^l = f^l(C_x) || f^l(C_y)
\label{eq:2c}
\end{align}
\end{subequations}
where $||$ denotes the concatenation operation performed along the channel dimension, $\alpha$ denotes a learnable parameter, \(Ma\) refers to the SSM module that models global information within the sequence, $F^{-1}$ denotes a scanning order selected based on Eq.~\ref{eq:1b} and Eq.~\ref{eq:1c}, and $w(C_i)$ denotes the weight at position $C_i$.

\subsection{Reverse Selective State Guidance}\label{se: RSSG}
Motivated by the blurred boundary representations often observed in deep-sampled features~\cite{pang2025efficient}, the RSSG module is designed to provide effective geometric structure guidance during upsampling. As illustrated in Fig.~\ref{fig: overview} (c), it integrates reverse guidance theory with the state-space mechanism to enhance the model’s ability to perceive geometric boundaries, thereby improving its focus on complex vascular topologies.


Specifically, each RSSG module takes three inputs: the shallow edge feature $F_{edge}$ extracted by CBAM, the deep semantic feature $F_i'$ obtained from the corresponding downsampling stage, and the reverse prediction $R_i$ from the preceding decoder block. As shown in Eq.~\ref{eq: rssg}, guided by the detailed spatial information from shallow features~\cite{du2022icgnet}, the RSSG module constructs a unified feature state-space by integrating decoder features and semantic base representations, which enhances the decoding process and promotes more accurate segmentation around geometrically complex boundaries.

\begin{equation}
\begin{aligned}
r = (-1 \times & Sigmoid(R_i) + 1) \times F_i' ; \\
f_c &= mmc (F_{edge} || r) ; \\
&f_m =  Ma(f_c) ; \\
&f_l =  mlp(f_m) ; \\
out_r = & f_l \times f_m \times f_c + F_i' . 
\end{aligned} 
\label{eq: rssg}
\end{equation}
where $out_r$ denotes the output of the RSSG module, $mlp$ represents a multilayer perceptron, $mmc(\cdot)$ indicates an MMC layer with a $3\times3$ kernel, and $\text{Sigmoid}$ denotes the sigmoid activation function~\cite{ren2007potential}.

\section{Experiments}

\subsection{Datasets}
In this study, we utilized two well-adapted datasets with vessel annotations: \textbf{DRIVE} and \textbf{STARE}. \\

\textbf{DRIVE}~\cite{staal2004ridge}: This dataset consists of 40 fundus images (565 × 584 pixels) collected from a diabetic retinopathy screening program in the Netherlands. The images are evenly split into training and test sets. Each image includes a field-of-view mask and corresponding vessel annotations. For the test set, two sets of vessel labels are provided: one as the gold standard and one from a second human observer. To facilitate model training, all images are resized to 608 × 608 pixels.

\textbf{STARE}~\cite{hoover2000locating}: This dataset contains 20 fundus images, each with a resolution of 605 × 700 pixels. Half of the cases present retinal vascular abnormalities. All images are annotated by clinical experts. Fifty percent of the images are allocated for training, and the remainder for testing. To facilitate model training, all images are resized to 704 × 704 pixels.

\subsection{Implementation Details}
Our model is implemented with PyTorch 2.0.0 and trained on NVIDIA Tesla V100S-PCIE-32GB GPUs with CUDA 12.4 support. We train the model for 500 epochs using the AdamW optimizer. A batch size of 5 is used for training and validation on the DRIVE dataset, and a batch size of 2 is used on the STARE dataset. The initial learning rate is set to 0.001, with a linear warm-up during the first two epochs, followed by a cosine annealing schedule that gradually decays the learning rate to a minimum of 1e-7. A weight decay of 0.05 is applied at the beginning of training and reduced to 0.04 by the final stage. 

\begin{table*}[t]
\centering
\renewcommand{\arraystretch}{1.1}
\setlength{\tabcolsep}{17pt}
\normalsize
\caption{Performance comparison of different methods on \textbf{DRIVE} and \textbf{STARE} datasets.}
\label{tab: performance}
\begin{tabular}{llccccc}
\hline
\textbf{Dataset} & \textbf{Architecture} & \textbf{ACC (\%)} & \textbf{Se (\%)} & \textbf{Sp (\%)} & \textbf{F1 (\%)} \\
\hline\hline
\multirow{11}{*}{DRIVE} 
& U-Net~\cite{ronneberger2015u}         & 95.56 & 75.56 & 97.30 & 79.97 \\
& DE-DCGCN-EE~\cite{li2022dual}         & 97.05 & 83.59 & 98.26 & 82.88 \\
& GT-DLA-dsHFF~\cite{9863763}           & 97.03 & 83.55 & 98.27 & 82.57 \\
& TP-Net~\cite{10018415}               & 96.29 & 87.49 & 97.58 & 85.69 \\
& BCU-Net~\cite{ZHANG2023106960}       & 96.62 & 82.38 & 98.00 & 80.89 \\
& Wave-Net~\cite{LIU2023106341}        & 95.61 & 81.64 & 97.64 & 82.54 \\
& CoVi-Net~\cite{JIANG2024108047}      & 96.98 & 83.47 & 98.30 & 87.48 \\
& HRD-Net~\cite{LIU2024108295}         & 97.04 & 83.71 & \underline{98.33} & 83.12 \\
& PA-Net~\cite{luo2025pa}              & 95.82 & 82.84 & 98.07 & 83.93 \\
& MSPDD-Net~\cite{li2025mspdd}         & \underline{97.45} & \underline{87.51} & 98.21 & \underline{87.95} \\
 \rowcolor[HTML]{EFEFEF} &MM-UNet (ours) & \textbf{98.27} & \textbf{89.33} & \textbf{99.08} & \textbf{89.59} \\
\hline

\multirow{11}{*}{STARE} 
& U-Net~\cite{ronneberger2015u}         & 96.17 & 81.67 & 98.33 & 81.12 \\
& DE-DCGCN-EE~\cite{li2022dual}         & 97.51 & 84.05 & 98.61 & 83.63 \\
& GT-DLA-dsHFF~\cite{9863763}           & 97.60 & 84.80 & 98.64 & 86.55 \\
& TP-Net~\cite{10018415}               & 97.24 & 88.52 & 98.20 & 86.75 \\
& BCU-Net~\cite{ZHANG2023106960}       & 97.01 & 85.00 & 98.07 & 82.23 \\
& Wave-Net~\cite{LIU2023106341}        & 96.41 & 79.02 & 98.36 & 81.40 \\
& CoVi-Net~\cite{JIANG2024108047}      & 97.61 & 83.05 & 98.87 & 90.31 \\
& HRD-Net~\cite{LIU2024108295}         & 97.55 & 84.59 & 98.62 & 83.57 \\
& PA-Net~\cite{luo2025pa}              & 97.09 & 88.13 & 98.05 & 85.61 \\
& MSPDD-Net~\cite{li2025mspdd}         & \underline{97.76} & \underline{88.75} & \underline{98.91} & \underline{90.52} \\
\rowcolor[HTML]{EFEFEF} & MM-UNet (ours) & \textbf{98.81} & \textbf{91.77} & \textbf{99.36} & \textbf{91.77} \\
\hline
\end{tabular}

\vspace{6pt}
\label{tab1}
\end{table*}

\subsection{Performance}
To comprehensively assess the effectiveness of our proposed MM-UNet framework, we conduct performance comparisons against a broad selection of state-of-the-art retinal vessel segmentation methods on two widely adopted benchmark datasets: DRIVE and STARE. These methods can be systematically categorized into four groups:

\begin{itemize}
    \item \textbf{Traditional Convolutional Segmentation Methods}: U-shaped Convolutional Network (U-Net)~\cite{ronneberger2015u}.
    
    \item \textbf{Graph and Multi-scale Convolutional Models}: Dual Encoder-based Dynamic-channel Graph Convolutional Network with Edge Enhancement (DE-DCGCN-EE)~\cite{li2022dual}, Global Transformer and Dual Local Attention Network via Deep-Shallow Hierarchical Feature Fusion (GT-DLA-dsHFF)~\cite{s22155826}, Bridge ConvNeXt U-Net (BCU-Net)~\cite{ZHANG2023106960}, and a Hybrid Architecture based on LPT and AFFM (PA-Net)
~\cite{luo2025pa}.
    
    \item \textbf{Transformer and Attention-enhanced Models}: Wave-Net~\cite{LIU2023106341} and Convolutional Vision Transformer Network (CoVi-Net)~\cite{JIANG2024108047}.
    
    \item \textbf{State-space and Dual-decoder Architectures}: Mamba Semantic Perception Dual-decoding Network (MSPDD-Net)~\cite{li2025mspdd}, High Resolution based on Deformable Convolution v3 (HRD-Net)~\cite{LIU2024108295}, and Two-Path Network (TP-Net)~\cite{10018415}.
\end{itemize}

All models are implemented and evaluated under identical experimental conditions to ensure fairness and reproducibility in comparison.

As shown in Table \ref{tab: performance}, MM-UNet achieves leading performance across all key evaluation metrics, including accuracy (ACC), sensitivity (Se), specificity (Sp), and F1-score~(F1), on both the DRIVE and STARE datasets. In the DRIVE dataset, MM-UNet reaches an F1-score of 89.59\% and sensitivity of 89.33\%, outperforming the best baseline, MSPDD-Net, which scores 87.95\% and 87.51\%, respectively.  Additionally, MM-UNet attains 98.27\% in ACC and 99.08\% in Sp. On the STARE dataset, MM-UNet again leads with 91.77\% in both F1-score and sensitivity, outperforming MSPDD-Net’s F1 of 90.52\% and Se of 88.75\%.

Specifically, although MSPDD-Net achieves strong ACC and Sp on the STARE dataset, its static context modeling limits adaptability to local structural variations. In contrast, MM-UNet employs MMC layers to enable dynamic, sub-pixel-level vessel perception, thereby enhancing continuity in sparse regions. On the other hand, while graph-based models such as GT-DLA-dsHFF and DE-DCGCN-EE offer improved global context representation, their relatively weak boundary localization leads to moderate performance in F1 and Se. The RSSG module in MM-UNet addresses this limitation by incorporating reverse-guided state-space modeling, which strengthens geometric boundary perception and results in sharper vessel delineation. Overall, the synergy between MMC and RSSG endows MM-UNet with superior structural accuracy and generalization capability, delivering consistent performance gains across retinal vessel segmentation datasets and evaluation metrics.

\begin{table*}[htbp]

\centering
\renewcommand{\arraystretch}{1.2}
\setlength{\tabcolsep}{17pt}
\normalsize
\caption{Ablation study on \textbf{DRIVE} and \textbf{STARE} datasets. The color scheme of this table is the same as that of Table \ref{tab1}.}
\label{tab: Ablation study}

\begin{tabular}{llcccc}
\hline
\textbf{Dataset} & \textbf{Architecture} & \textbf{ACC (\%)} & \textbf{Se (\%)} & \textbf{Sp (\%)} & \textbf{F1 (\%)} \\
\hline\hline
\multirow{3}{*}{DRIVE} 
& w/o MMC         & 97.70 & 85.94 & 98.77 & 86.21 \\
& w/o RSSG        & 96.91 & 80.99 & 98.36 & 81.41 \\
\rowcolor[HTML]{EFEFEF} &{\textbf{MM-UNet}} & \textbf{98.27} & \textbf{89.33} & \textbf{99.08} & \textbf{89.59} \\
\hline

\multirow{3}{*}{STARE} 
& w/o MMC         & 97.70 & 85.94 & 98.77 & 86.21 \\
& w/o RSSG        & 96.91 & 80.99 & 98.36 & 81.41 \\

\rowcolor[HTML]{EFEFEF} & \textbf{MM-UNet} & \textbf{98.81} & \textbf{91.77} & \textbf{99.36} & \textbf{91.77} \\
\hline
\label{tab: table1}
\end{tabular}

\end{table*}

\begin{figure*}[!t]
    \centering      
    \includegraphics[width=\textwidth]{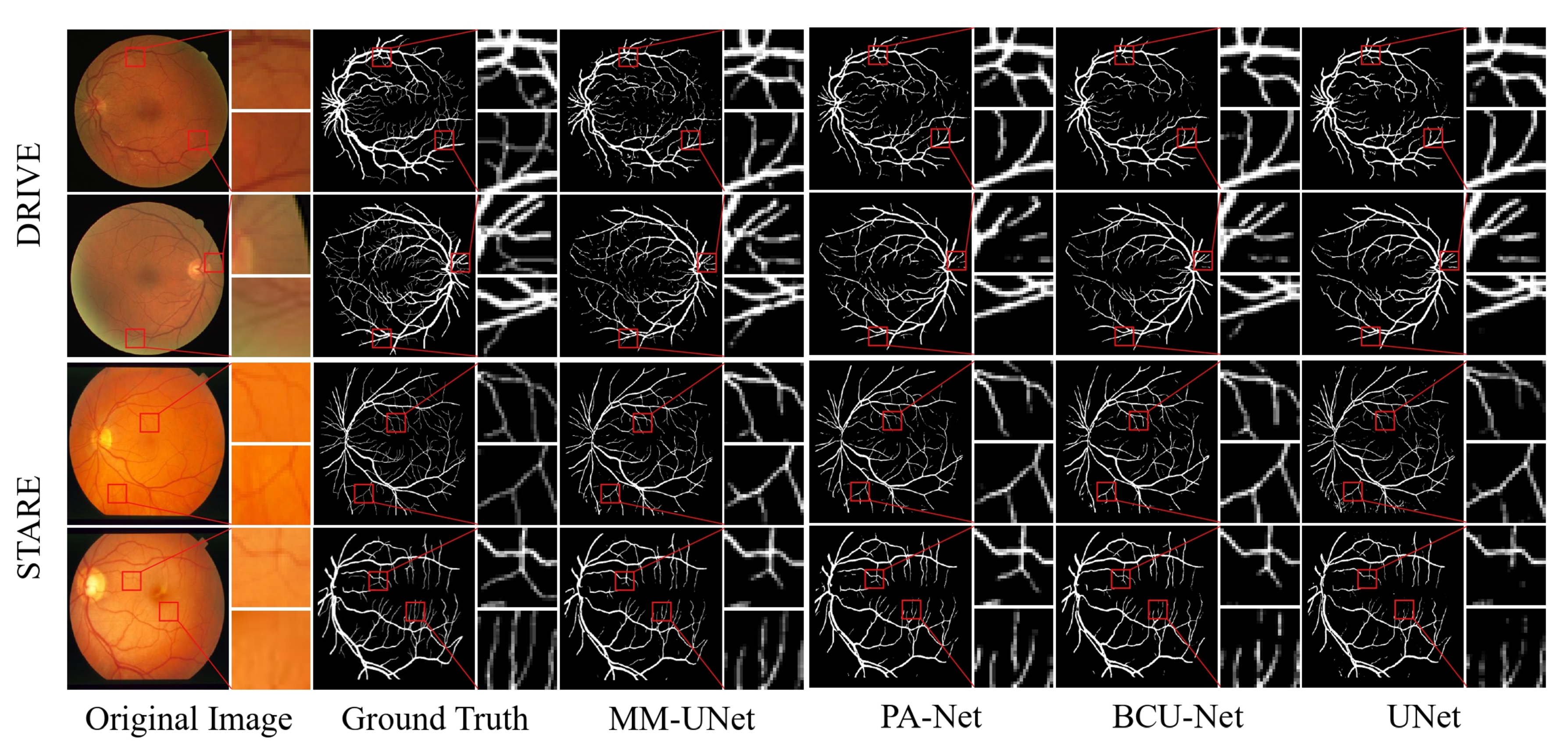}
    \caption{Visual comparisons of proposed MM-UNet and other SOTA methods.}
    \label{fig: fig3}
    \vspace{6pt}
\end{figure*}

\begin{figure*}[!t]
    \centering      
    \includegraphics[width=0.9\textwidth]{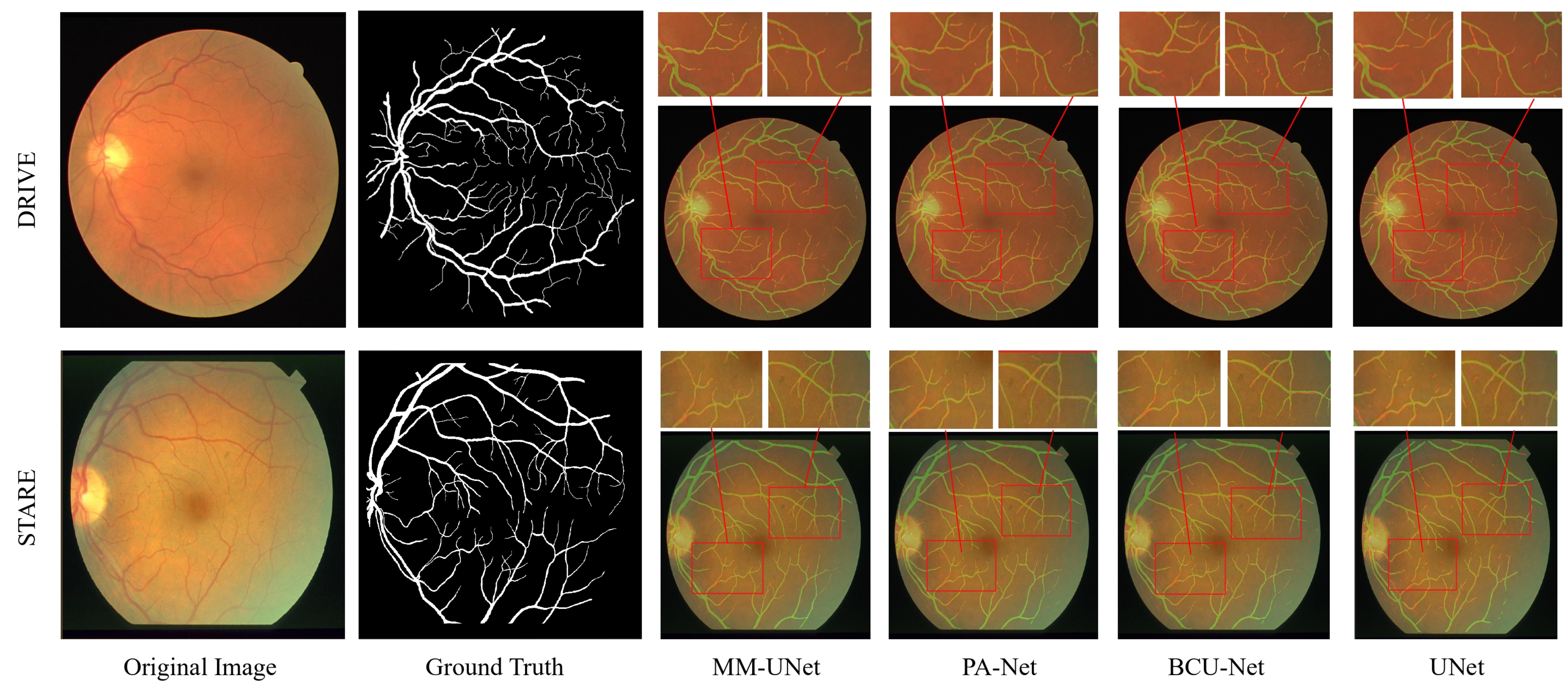}
    \caption{Error map of proposed MM-UNet and other SOTA methods.}
    \label{fig: fig4}
    \vspace{6pt}
\end{figure*}

\subsection{Ablation Study}
The effectiveness of the core components in MM-UNet is assessed through a comprehensive ablation study, wherein the MMC layers and the RSSG modules are individually removed.

As shown in Table \ref{tab: Ablation study}, a marked decline in segmentation performance is observed across all evaluation metrics when the MMC layers are replaced with conventional two-dimensional convolutional layers using identical hyperparameters (w/o MMC). Specifically, ACC drops to 96.91\%, Se to 80.99\%, and the F1 to 81.41\%. This degradation highlights the critical role of MMC layers, which integrates morph convolution with state-space modeling to enable dynamic, sub-pixel level feature sampling. These layers are particularly effective in representing the narrow, tortuous, and branching patterns of retinal vessels—especially at terminal segments—where traditional convolutions often fail to capture such intricate anatomical details. Likewise, Table \ref{tab: Ablation study} validates that removing the RSSG modules~(w/o RSSG) results in diminished performance, particularly along vessel boundaries. This module enhances structural delineation by fusing shallow edge features with deep semantic representations via a reverse-guided state-space mechanism. Its absence compromises the model’s ability to perceive and localize vessel contours accurately, leading to less coherent and less precise segmentation results.

In contrast, the full MM-UNet architecture, which incorporates both Morph Mamba Convolution and Reverse Selective State Guidance, achieves notable performance gains—yielding an accuracy of 99.82\%, sensitivity of 98.78\%, and an F1-score of 98.84\%. These findings affirm the indispensability of both modules in driving robust and high-fidelity segmentation, particularly in preserving fine structural details and ensuring boundary continuity.

\subsection{Visual Comparisons}
The performance visual results provide a more intuitive demonstration of the superior performance of MM-UNet in RVS. Furthermore, the presented error maps effectively highlight its capability in accurately capturing complex vascular topologies:
\subsubsection{Performance}
Fig. \ref{fig: fig3} visually compares the segmentation results of different methods. On the DRIVE and STARE datasets, our MM‑UNet accurately delineates the boundaries of all retinal vessels and demonstrates superior consistency compared to other SOTA approaches.
\subsubsection{Error Map}
Fig.~\ref{fig: fig4} intuitively illustrates the performance of various methods in segmenting vascular branches. The selected error maps from the DRIVE and STARE datasets, which include both bright and dark regions, further demonstrate that our MM-UNet not only achieves higher accuracy in vascular branch segmentation but also exhibits strong generalization capability under varying illumination conditions.

\section{Conclusion}
In this study, we proposed MM-UNet, a novel and robust framework tailored for retinal vessel segmentation, which effectively addresses the inherent challenges of complex tubular morphology, ambiguous boundary delineation, and multi-scale structural variability in fundus images. The proposed architecture incorporates two key innovations: (1) Morph Mamba Convolution layers, which replace conventional pointwise feature sampling in U-shaped networks by integrating dynamic morph convolution with selective state-space modeling, thereby improving the network’s sensitivity to topological continuity and thin tubular structures; (2) Reverse Selective State Guidance modules, which embed reverse attention mechanisms within a hierarchical state-space-guided fusion strategy to reinforce boundary-level discrimination and efficient cross-scale information propagation. Extensive experiments conducted on the DRIVE and STARE datasets demonstrate the superiority of MM-UNet over current state-of-the-art methods. MM-UNet achieves F1-scores of 89.59\% on DRIVE and 91.77\% on STARE, with relative improvements compared to the second-best performing models. These results validate the effectiveness, generalizability, and practical potential of our framework across diverse image acquisition conditions and retinal vessel distributions.

\section{Acknowledgement}
This work was supported in part by the National Key Research and Development Program of China (2024YFF1206600, 2022YFE0112200); in part by the National Natural Science Foundation of China (U21A20520, 62325204, 62502161, 62102153, 62272326, 62172112); in part by the Key-Area Research and Development Program of Guangzhou City (2023B01J1001, 2023B01J0002); in part by the Science and
Technology Project of Guangdong Province under Grant 2022A0505050014;
in part by the Key-Area Research and Development Program of Guangzhou
City under Grant 202206030009; in part by the Natural Science Foundation
of Guangdong Province of China under Grant 2022A1515011162 and
Grant 2023A1515012894; in part by the Guangdong Natural Science Funds for
Distinguished Young Scholar under Grant 2023B1515020097.



\bibliographystyle{splncs04}
\bibliography{cite}

\vspace{12pt}

\end{document}